\documentclass[a4paper]{article}

\usepackage{INTERSPEECH2018}

\usepackage{cite}

\title{Articulatory Features for ASR of Pathological Speech}
\name{Emre Y\i lmaz$^{1,2}$, Vikramjit Mitra$^3$, Chris Bartels$^4$ and Horacio Franco$^4$\thanks{The second author is currently working with Apple Inc. This research is funded by the NWO Project 314-99-101 (CHASING) and NWO Project 314-99-119 (Frisian Audio Mining Enterprise).}}
\address{
  $^1$CLS/CLST, Radboud University, Nijmegen, Netherlands\\
  $^2$ Dept. of Electrical and Computer Engineering, National University of Singapore, Singapore \\
  $^3$University of Maryland, College Park, MD, USA\\
  $^4$STAR Lab, SRI International, Menlo Park, CA, USA}
\email{e.yilmaz@let.ru.nl, vmitra@umd.edu, \\\{chris.bartels,horacio.franco\}@sri.com}

\begin{document}

\maketitle
\begin{abstract}
In this work, we investigate the joint use of articulatory and acoustic features for automatic speech recognition (ASR) of pathological speech. Despite long-lasting efforts to build speaker- and text-independent ASR systems for people with dysarthria, the performance of state-of-the-art systems is still considerably lower on this type of speech than on normal speech. The most prominent reason for the inferior performance is the high variability in pathological speech that is characterized by the spectrotemporal deviations caused by articulatory impairments due to various etiologies. To cope with this high variation, we propose to use speech representations which utilize articulatory information together with the acoustic properties. A designated acoustic model, namely a fused-feature-map convolutional neural network (fCNN), which performs frequency convolution on acoustic features and time convolution on articulatory features is trained and tested on a Dutch and a Flemish pathological speech corpus. The ASR performance of fCNN-based ASR system using joint features is compared to other neural network architectures such conventional CNNs and time-frequency convolutional networks (TFCNNs) in several training scenarios.
\end{abstract}

\section{Introduction}

Speech disorders causing deviations in articulation lead to decreased speech intelligibility and communication impairment \cite{kent2003}. Recent developments show that therapy can be provided by employing computer-assisted speech training systems \cite{beijer2011}. According to the outcomes of the efficacy tests presented in \cite{beijer2014}, user satisfaction towards such a system appears to be quite high. However, most of these systems are not yet capable of automatically detecting problems at the level of individual speech sounds, which are known to have an impact on speech intelligibility \cite{debodt2002,yunusova2005,nuffelen2009,popovici2012,ganzeboom2016}. 
Our goal is to develop robust automatic speech recognition (ASR) systems for pathological speech and incorporate the ASR technology to detect these problems.

Training robust acoustic models to capture the within- and between-speaker variation in dysarthric speech is generally not feasible due to the limited size and structure of existing pathological speech databases. The number of recordings in dysarthric speech databases is much smaller compared to that in normal speech databases. Despite long-lasting efforts to build speaker- and text-independent ASR systems for people with dysarthria, the performance of state-of-the-art systems is still considerably lower on this type of speech than on normal speech \cite{sanders2002,rudzicz2007,caballero2009,mengistu2011,seong2012,christensen2012,shahamiri2014,takashima2015,lee2016}.

In previous work \cite{yilmaz2016_5}, we described a solution to train a better DNN-hidden Markov model (HMM) system for the Dutch language, a language that has fewer speakers and resources compared to English. In particular, we investigated combining non-dysarthric speech data from different varieties of the Dutch language to train more reliable acoustic models for a DNN-HMM ASR system. This work was conducted in the framework of the CHASING project\footnote{http://hstrik.ruhosting.nl/chasing/}, in which a serious game employing ASR is being developed to provide additional speech therapy to dysarthric patients \cite{ganzeboom2016_2}. Moreover, we created a 6-hour Dutch dysarthric speech database that had been collected in a previous project (EST) \cite{yilmaz2016_2} for training purposes and investigate the impact of multi-stage DNN training for pathological speech~\cite{yilmaz2017}.

Using articulatory features (AF) together with acoustic features has been investigated and shown to be beneficial in the ASR of normal speech, e.g. \cite{zlokarnik1995,wrench2000,stephenson2000,markov2006,mitra2012,badino2016}. 
A subset of these approaches learn a mapping between acoustic and articulatory spaces for the speech inversion, and use the learned articulatory information in an ASR system for improved representation of speech in a high-dimensional feature space. Rudzicz \cite{rudzicz2011} tried using AF together with conventional acoustic features for phone classification experiments on dysarhtric speech. More recently, \cite{mitra2017} has proposed the use of convolutional neural networks (CNN) for learning speaker independent articulatory models for mapping acoustic features to the corresponding articulatory space. Later, a novel acoustic model designed to integrate the AF together with the acoustic features has been proposed \cite{mitra2017_2}.

In this work, we investigate the joint use of articulatory and acoustic features for the ASR of pathological speech. Specifically, we explore the use of vocal tract constriction variables (TVs) and standard filterbank features as input to fused-feature-map CNN (fCNN) acoustic models as described in~\cite{mitra2017_2}. Incorporating articulatory information in the features for the ASR of pathological speech is expected to increase the robustness against increased spectrotemporal deviations due to reduced articulatory capabilities of the speakers.~To investigate the impact of articulatory knowledge for ASR of pathological speech, we train fCNN acoustic models using the concatenated acoustic and articulatory features and evaluate the ASR performance on two different pathological speech corpora with varying levels of dysarthria. The performance of this system is compared to other NN-based acoustic models such as conventional deep neural networks (DNN), CNN and time-frequency CNNs (TFCNN).

The rest of the paper is organized as follows. Section \ref{sec:scs} explains the selection of various speech corpora for the proposed training scheme. Section \ref{sec:af} describes how the AFs are extracted and used in an ASR system. Section \ref{sec:acmod} summarizes the acoustic models used in this work. The experimental setup is described in Section \ref{sec:expset} and the recognition results are presented in Section \ref{sec:res}. Section \ref{sec:conc} concludes the paper.
\vspace{-0.2cm}
\section{Speech corpora selection}
\label{sec:scs}
Given the limited availability of dysarthric speech data, we investigate to what extent already existing databases of Dutch normal speech can be employed to train NN-based acoustic models and optimize their performance on dysarthric speech. There have been multiple Dutch-Flemish speech data collection efforts~\cite{cgn,jasmin} which facilitate the integration of both Dutch and Flemish data in the present research. For training purposes, we used the CGN corpus~\cite{cgn}, which contains representative collections of contemporary standard Dutch as spoken by adults in the Netherlands and Flanders. The CGN components with read speech, spontaneous conversations, interviews and discussions are used for acoustic model training. The duration of the normal Dutch (NL) and Flemish (FL) speech data used training is 255 and 186.5 hours respectively. The combined training data (FL+NL) contains 441.5 hours in total.

The EST Dutch dysarthric speech database~\cite{yilmaz2016_2} contains dysarthric speech from ten patients with Parkinson's Disease (PD), four patients who have had a Cerebral Vascular Accident (CVA), one patient who suffered Traumatic Brain Injury (TBI) and one patient having dysarthria due to a birth defect. Based on the meta-information, the age of the speakers is in the range of 34 to 75 years with a median of 66.5 years. The level of dysarthria varies from mild to moderate. The dysarthric speech collection for this database was achieved in several experimental contexts. The speech tasks presented to the patients in these contexts consist of numerous word and sentence lists with varying linguistic complexity. The database includes 12 Semantically Unpredictable Sentences (SUSs) with 6- and 13-word declarative sentences, 12 6-word interrogative sentences, 13 Plomp and Mimpen sentences, 5 short texts, 30 sentences with /t/, /p/ and /k/ in initial position and unstressed syllable, 15 sentences with /a/, /e/ and /o/ in unstressed syllables, production of 3 individual vowels /a/, /e/ and /o/, 15 bisyllabic words with /t/, /p/ and /k/ in initial position and unstressed syllable and 25 words with alternating vowel-consonant composition (CVC, CVCVCC, etc.).

For testing purposes, we firstly use the sentence reading tasks of the CHASING01 Dutch dysarthric speech database~\cite{yilmaz2017}. This database contains speech of 5 patients who participated in speech training experiments and were tested at 6 different times during the treatment. For each set of audio files, the following material was collected: 12 SUSs, 30 /p/, /t/, /k/ sentences in which the first syllable of the last word is unstressed and starts with /p/, /t/ or /k/, 15 vowel sentences with the vowels /a/,/e/ and /o/ in stressed syllables, appeltaarttekst (\textit{apple cake recipe}) in 5 parts. Utterances that deviated from the reference text due to pronunciation errors (e.g. restarts, repeats, hesitations, etc.) were removed. After this subselection, the utterances from 3 male patients remained and were included in the test set. These speakers are 67, 62 and 59 years old, two of them having PD and the third having had a CVA.

All sentence reading tasks with annotations from the COPAS pathological speech corpus~\cite{middagphd}, namely 2 isolated sentence reading tasks, 11 text passages with reading level difficulty of AVI 7 and 8 and Text Marloes, are also included as a second test set. The COPAS corpus contains recordings from 122 Flemish normal speakers and 197 Flemish speakers with speech disorders such as dysarthria, cleft, voice disorders, laryngectomy and glossectomy. The dysarthric speech component contains recordings from 75 Flemish patients affected by Parkinson's disease, traumatic brain injury, cerebrovascular accident and multiple sclerosis who exhibit dysarthria at different levels of severity. Containing more speakers with more diverse etiologies, performing ASR on this corpus is found to more challenging compared to the CHASING01 dysarthric speech database (c.f. the ASR results in~\cite{yilmaz2016_5} and~\cite{yilmaz2017}).
\vspace{-0.2cm}
\section{Extracting Articulatory Features}
\label{sec:af}
The task of estimating the articulatory trajectories (in this case, the vocal tract constriction variables (TVs)) from the speech signal is commonly known as speech-to-articulatory inversion or simply speech-inversion. TVs~\cite{mitra2011,browman1992} are continuous time functions that specify the shape of the vocal tract in terms of constriction degree and location of the constrictors. During speech-inversion, the acoustic features extracted from the speech signal are used to predict the articulatory trajectories, where the inverse mapping is learned by using a parallel corpus containing acoustic and articulatory pairs. The task of speech-inversion is a well-known, ill-posed inverse transform problem, which suffers from both the non-linearity and non-unique nature of the inverse transform~\cite{richmond2001,vikramphd}.

The articulatory dataset used to train the speech-inversion systems consists of synthetic speech with simultaneous tract variable trajectories. We used the Haskins Laboratories' Task Dynamic model (TADA)~\cite{nam2004} along with HLsyn~\cite{manson2002} to generate a synthetic English isolated word speech corpus along with TVs. Altogether 534\,322 audio samples were generated (approximately 450 h of speech), out of which 88\% of the data was used as the training set, 2\% was used as the cross-validation set, and the remaining 10\% was used as the test set. We further added fourteen different noise types (such as babble, factory noise, traffic noise, highway noise, crowd noise, etc.) to each of the synthetic acoustic waveforms with a signal-to-noise ratio (SNR) between 10 and 80 dB. We combined this noise-added data with the clean data, and the resulting combined dataset is used for learning a CNN-based speech inversion system. For further details, we refer the reader to~\cite{mitra2017}.

In this work, we use speech subband amplitude modulation features such as normalized modulation coefficients (NMCs)~\cite{mitra2014}. NMCs are noise-robust acoustic features obtained from tracking the amplitude modulations (AM) of filtered subband speech signals in the time domain. The features are Z-normalized before being used to train the CNN models. Further, the input features are contextualized by splicing multiple frames. Given the linguistic similarity between English and Dutch, we assume that the speech inversion model trained on English speech would give a reasonably accurate acoustic-to-articulatory mapping in Dutch. For a detailed comparison of the articulatory setting in Dutch and English, please see Section 21 of~\cite{collins1996}.

\begin{figure*}
  \includegraphics[width=6.8in, trim={1cm 3cm 0 5cm}]{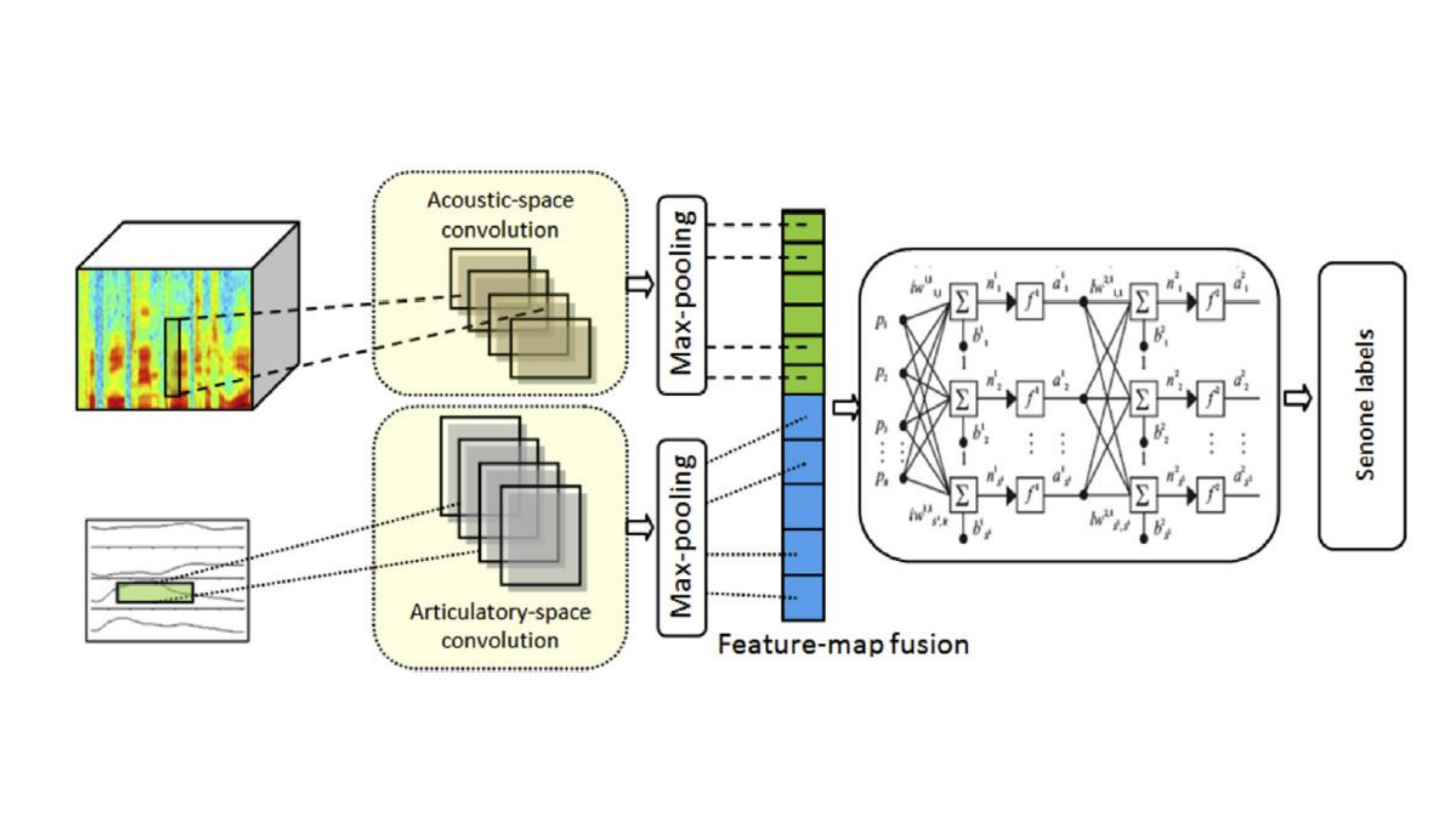}
  \caption{A fused-feature-map convolutional neural network (fCNN)~\cite{mitra2017_2}}
  \label{fig:fcnn}
  \vspace{-0.5cm}
\end{figure*}
\vspace{-0.2cm}
\section{Acoustic Models}
\label{sec:acmod}

The acoustic and articulatory (concatenated) features are fed to a fused-feature-map convolutional neural network (fCNN) which is illustrated in Figure \ref{fig:fcnn}. This architecture uses two types of convolutional layers. The first convolutional layer operates on the acoustic features, which are the filterbank energy features, and performs convolution across frequency. The other convolutional layer operates on AFs, which are the TV trajectories, and performs convolution across time. The output of the max-pooling layers are fed to a single NN after performing feature-map fusion.

Time-frequency convolutional nets (TFCNN) are also a suitable candidate for the acoustic modeling of dysarthric speech. Performing convolution both on time and frequency axes, they exhibit increased robustness against the spectrotemporal deviations due to background noise~\cite{mitra2015}. In the scope of this work, we use these models as one of the alternative ASR systems without delving into their advantages compared to other NN architectures. This investigation remains as a future work. We further train baseline DNN and CNN models using filterbank features as these architectures are found to provide worse recognition performance using the concatenated features~\cite{mitra2017_2}.
\vspace{-0.2cm}
\section{Experimental Setup}
\label{sec:expset}

\subsection{Database details}
\label{ssec:database}

The CGN components with read speech, spontaneous conversations, interviews and discussions were used for acoustic model training. The duration of the normal Flemish (FL) and northern Dutch (NL) speech data used for training is 186.5 and 255 hours, respectively.

The EST Dutch dysarthric speech database (Dys. NL) contains 6 hours and 16 minutes of dysarthric speech material from 16 speakers~\cite{yilmaz2016_2}. The speech segments with pronunciation errors (e.g. restarts, repeats, hesitations, etc.) were excluded from the training set to maintain integrity of the results on ASR performance evaluation. Additionally, the segments including a single word and pseudoword were also excluded, since the sentence reading tasks are more relevant in our project context. The total duration of the dysarthric speech data eventually selected for training is 4 hours and 47 minutes.

For testing purposes, we use two databases: (1) The CHASING01 Dutch dysarthric speech data which contains 721 utterances (in total 6231 words) spoken by 3 dysarthric speakers with a total duration of 55 minutes, (2) The Flemish COPAS database which contains 212 different sentence tasks uttered by 103 dysarthric and 82 normal speakers. The sentence tasks uttered in the Flemish corpus by normal speakers (SentNor) and speakers with disorders (SentDys) consists of 1918 (15,149) and 1034 (8287) sentences (words) with a total duration of 1.5 and 1 hour, respectively.

\subsection{Implementation Details}
\label{ssec:impdet}

We use CNNs for training speech inversion models, where contextualized (spliced) acoustic features in the form of NMCs are used as input, and the TV trajectories were used as the targets. The network parameters and the splicing window were optimized by using a held-out development set. The convolution layer of the CNN had 200 filters, where max-pooling was performed over three samples. The CNN has three fully connected hidden layers with 2048 neurons in each layer. For further details, we refer the reader to~\cite{mitra2017}.

For ASR experiments, a conventional context dependent GMM-HMM system with 40k Gaussians was trained on the 39-dimensional MFCC features including the deltas and delta-deltas. We also trained a GMM-HMM system on the LDA-MLLT features, followed by training models with speaker adaptive training using FMLLR features. This system was used to obtain the state alignments required for NN training. The input features to the acoustic models are formed by using a context window of 17 frames (8 frames on either side of the current frame).

The acoustic models were trained by using cross-entropy (CE) on the alignments from the GMM-HMM system. The 40-dimensional log-mel filterbank (FB) features with the deltas and delta-deltas are used as acoustic features which are extracted using the Kaldi~\cite{kaldi} toolkit. The NN models are implemented in Theano. The NNs trained on dysarthric Dutch training data has 4 hidden layers, with 1024 nodes per hidden layer. The NNs trained on normal Dutch and Flemish data has 6 hidden layers, with 2048 nodes per hidden layer, and the output layer included as many nodes as the number of CD states for the given dataset. The networks were trained by using an initial four iterations with a constant learning rate of 0.008, followed by learning-rate halving based on cross validation error decrease. Training stopped when no further significant reduction in cross-validation error was noted or when cross-validation error started to increase. Back-propagation was performed using stochastic gradient descent with a mini-batch of 256 training examples. All ASR systems use the Kaldi decoder.

For CNN, TFCNN and fCNN, the acoustic space is learned using a 200 convolutional filters of size 8 were used in the convolutional layer, and the pooling size was set to 3 without overlap. For fCNN, the articulatory space is learned by using a time-convolution layer that contains 75 filters, followed by max-pooling over 5 samples. Further implementation details about the NN parameters are available in~\cite{mitra2017_2}.

\begin{table}[!t]
\centering
\caption{Word error rates in \% obtained on the Dutch test set using different acoustic models}
\vspace{-0.3cm}
\begin{tabular}{| c | c | c | c |}
\hline
AM        	&	Features	& Train. Data 	&  WER (\%)\\
\hline \hline
DNN         &   FB		& Dys. NL     		&	22.9    \\
\hline
CNN         &   FB		& Dys. NL    		&   21.1	\\
\hline
TFCNN       &   FB		& Dys. NL     		&   20.3	\\
\hline \hline
fCNN        &   FB + TV & Dys. NL  &  \bf{19.1}  \\
\hline \hline
DNN         &   FB	    & Nor. NL     		&	15.0	\\
\hline
CNN         &   FB		& Nor. NL    		&  	14.9	\\
\hline
TFCNN       &   FB		& Nor. NL     		&   \bf{14.1}	\\
\hline \hline
fCNN        &   FB + TV & Nor. NL  & 15.0	\\
\hline
\end{tabular}
\label{tab:res_nl}
\vspace{-0.6cm}
\end{table}

\subsection{ASR experiments}
\label{ssec:asrexp}

We use two training setups for each test set during the ASR experiments. For the Dutch test data, the ASR system is either trained on normal or dysarthric Dutch speech. Training on combination of these databases has yielded very similar results to the system trained only the normal Dutch data in the pilot experiments. Therefore, we do not consider this training setup in this paper. 

For Flemish test data, we use normal Flemish and Dutch speech due to lack of training material in this language variety. In the first setting, we only use normal Flemish speech to train acoustic models, while both normal Flemish and Dutch speech is used in the second setting motivated by the improvements reported in~\cite{yilmaz2016_2}. The recognition performance of all ASR systems is quantified using the Word Error Rate (WER).
\vspace{-0.1cm}
\section{Results and Discussion}
\label{sec:res}

The ASR results obtained on the Dutch test set are presented in Table \ref{tab:res_nl}. The WERs provided by different acoustic models trained on the dysarthric Dutch speech are given in the upper panel of this table. The best ASR performance of each panel is marked in bold. Only using 6 hours of in-domain speech, the designated ASR system using both the acoustic and articulatory features provides the best ASR performance with a WER of 19.1\%. The CNN and TFCNN models trained on filterbank features provide a WER of 21.1\% and 20.3\% respectively. 

The ASR performance of the acoustic models trained on normal Dutch speech is given in the lower panel. In this scenario, the TFCNN model has the best performance with a WER of 14.1\%, while the other systems provide comparable recognition accuracies. In this training setting, in which we use large amount of mismatched data for the ASR of mild-to-moderate pathological speech, using articulatory information does not turn out to be bring further improvements compared to an ordinary CNN model.

We test the proposed recognition scheme on the Flemish corpus which contains speech data from much more dysarthric speakers (103 speakers compared to the 3 of the Dutch corpus). The ASR results obtained on the Flemish test set are presented in Table \ref{tab:res_vl}. In the Flemish test set, we also present the performance on the control data which contains similar sentence tasks uttered by normal speakers. In the first training scenario, we only use normal Flemish speech. The fCNN model provides a WER of 32.2\% which is considerably better than the 33.8\% of TFCNN and 33.5\% of CNN models. Consistent with~\cite{yilmaz2016_5}, when we add normal Dutch speech to the training data, we get a general improvement in the ASR performance on the Flemish test set. The fCNN model outperforms the other models with a WER of 29.0\%. 

\begin{table}[!t]
\centering
\caption{Word error rates in \% obtained on the Flemish test sets using different acoustic models}
\vspace{-0.3cm}
\addtolength{\tabcolsep}{-2.8pt}
\begin{tabular}{| c | c | c | c | c |}
\hline
AM        	&	Features	& Train. Data 	&  SentDys & SentNor    \\
\hline \hline
DNN         &   FB		& Nor. VL     		&	36.4	&    6.0	\\
\hline
CNN         &   FB		& Nor. VL    		&   33.5	&	 5.3	\\
\hline
TFCNN       &   FB		& Nor. VL     		&   33.8    &     5.3 \\
\hline \hline
fCNN        &   FB + TV & Nor. VL  &  \textbf{32.2}	&	5.0	 \\
\hline \hline
DNN         &   FB		& Nor. VL + Nor. NL 		&	32.1	& 5.5   \\
\hline
CNN         &   FB		& Nor. VL + Nor. NL 		&   30.1	& 4.9 	\\
\hline
TFCNN       &   FB		& Nor. VL + Nor. NL    		&   30.1	& 4.9 	\\
\hline \hline
fCNN        &   FB + TV & Nor. VL + Nor. NL &  \bf{29.0}	& 4.9 \\
\hline
\end{tabular}
\label{tab:res_vl}
\vspace{-0.6cm}
\end{table}

Even though there is still a large gap with the performance on the control data, using articulatory features with a designated NN architecture provides consistently improved ASR performance on the Flemish test set which contains speech from 103 dysarthric speakers. In general, these results demonstrate the potential of jointly using AFs and acoustic features against the spectrotemporal deviations in the pathological speech.
\vspace{-0.1cm}
\section{Conclusions}
\label{sec:conc}

In this work, we investigate incorporating articulatory and acoustic features jointly in the ASR of pathological speech. The ASR systems operating on this kind of speech suffers from the increased speech variation due to the poor articulation capabilities of the speakers. We explore the impact of using articulatory information in a ASR system by training various acoustic models in several scenarios and testing on a Dutch and a Flemish pathological speech corpus. The results demonstrate that using AF features brings improvements using a limited amount of in-domain training data. Moreover, we observed consistent improvements in the ASR performance in more challenging testing conditions with considerably higher number of speakers with a speech pathology originating from more diverse etiologies.
\vspace{-0.3cm}
\section{Acknowledgements}
We would like to thank Dimitra Vergyri and Helmer Strik for useful discussion, Aaron Lawson and Mitchell McLaren for the arrangements making this collaboration possible.
\bibliographystyle{IEEEtran}

\bibliography{refs}

\begin{thebibliography}{10}
\providecommand{\url}[1]{#1}
\csname url@samestyle\endcsname
\providecommand{\newblock}{\relax}
\providecommand{\bibinfo}[2]{#2}
\providecommand{\BIBentrySTDinterwordspacing}{\spaceskip=0pt\relax}
\providecommand{\BIBentryALTinterwordstretchfactor}{4}
\providecommand{\BIBentryALTinterwordspacing}{\spaceskip=\fontdimen2\font plus
\BIBentryALTinterwordstretchfactor\fontdimen3\font minus
  \fontdimen4\font\relax}
\providecommand{\BIBforeignlanguage}[2]{{%
\expandafter\ifx\csname l@#1\endcsname\relax
\typeout{** WARNING: IEEEtran.bst: No hyphenation pattern has been}%
\typeout{** loaded for the language `#1'. Using the pattern for}%
\typeout{** the default language instead.}%
\else
\language=\csname l@#1\endcsname
\fi
#2}}
\providecommand{\BIBdecl}{\relax}
\BIBdecl

\bibitem{kent2003}
R.~D. Kent and Y.~J. Kim, ``Toward an acoustic topology of motor speech
  disorders,'' \emph{Clin Linguist Phon}, vol.~17, no.~6, pp. 427--445, 2003.

\bibitem{beijer2011}
L.~J. Beijer and A.~C.~M. Rietveld, ``Potentials of telehealth devices for
  speech therapy in {Parkinson's} disease, diagnostics and rehabilitation of
  {Parkinson's} disease,'' \emph{InTech}, pp. 379--402, 2011.

\bibitem{beijer2014}
L.~J. Beijer, A.~C.~M. Rietveld, M.~B. Ruiter, and A.~C. Geurts, ``Preparing an
  {E-learning-based Speech Therapy (EST)} efficacy study: Identifying suitable
  outcome measures to detect within-subject changes of speech intelligibility
  in dysarthric speakers,'' \emph{Clinical Linguistics and Phonetics}, vol.~28,
  no.~12, pp. 927--950, 2014.

\bibitem{debodt2002}
M.~S. De~Bodt, H.~M. Hernandez-Diaz, and P.~H. Van De~Heyning,
  ``Intelligibility as a linear combination of dimensions in dysarthric
  speech,'' \emph{Journal of Communication Disorders}, vol.~35, no.~3, pp.
  283--292, 2002.

\bibitem{yunusova2005}
Y.~Yunusova, G.~Weismer, R.~D. Kent, and N.~M. Rusche, ``Breath-group
  intelligibility in dysarthria: characteristics and underlying correlates,''
  \emph{J Speech Lang Hear Res.}, vol.~48, no.~6, pp. 1294--1310, 2005.

\bibitem{nuffelen2009}
G.~Van~Nuffelen, C.~Middag, M.~De~Bodt, and J.-P. Martens, ``Speech
  technology-based assessment of phoneme intelligibility in dysarthria,''
  \emph{International Journal of Language \& Communication Disorders}, vol.~44,
  no.~5, pp. 716--730, 2009.

\bibitem{popovici2012}
D.~V. Popovici and C.~Buic\u{a}-Belciu, ``Professional challenges in
  computer-assisted speech therapy,'' \emph{Procedia - Social and Behavioral
  Sciences}, vol.~33, pp. 518 -- 522, 2012.

\bibitem{ganzeboom2016}
M.~Ganzeboom, M.~Bakker, C.~Cucchiarini, and H.~Strik, ``Intelligibility of
  disordered speech: Global and detailed scores,'' in \emph{Proc. INTERSPEECH},
  Sept. 2016, pp. 2503--2507.

\bibitem{sanders2002}
E.~Sanders, M.~B. Ruiter, L.~J. Beijer, and H.~Strik, ``Automatic recognition
  of {Dutch} dysarthric speech: a pilot study,'' in \emph{Proc. INTERSPEECH},
  2002, pp. 661--664.

\bibitem{rudzicz2007}
F.~Rudzicz, ``Comparing speaker-dependent and speaker-adaptive acoustic models
  for recognizing dysarthric speech,'' in \emph{Proc. of the 9th International
  ACM SIGACCESS Conference on Computers and Accessibility}, 2007, pp. 255--256.

\bibitem{caballero2009}
S.-O. Caballero-Morales and S.~J. Cox, ``Modelling errors in automatic speech
  recognition for dysarthric speakers,'' \emph{EURASIP J. Adv. Signal Process},
  pp. 1--14, Jan. 2009.

\bibitem{mengistu2011}
K.~T. Mengistu and F.~Rudzicz, ``Adapting acoustic and lexical models to
  dysarthric speech,'' in \emph{Proc. ICASSP}, may 2011, pp. 4924--4927.

\bibitem{seong2012}
W.~Seong, J.~Park, and H.~Kim, ``Dysarthric speech recognition error correction
  using weighted finite state transducers based on context-dependent
  pronunciation variation,'' in \emph{Computers Helping People with Special
  Needs}, ser. Lecture Notes in Computer Science, 2012, vol. 7383, pp.
  475--482.

\bibitem{christensen2012}
H.~Christensen, S.~Cunningham, C.~Fox, P.~Green, and T.~Hain, ``A comparative
  study of adaptive, automatic recognition of disordered speech.'' in
  \emph{INTERSPEECH}, 2012, pp. 1776--1779.

\bibitem{shahamiri2014}
S.~R. Shahamiri and S.~S.~B. Salim, ``Artificial neural networks as speech
  recognisers for dysarthric speech: Identifying the best-performing set of
  {MFCC} parameters and studying a speaker-independent approach,''
  \emph{Advanced Engineering Informatics}, vol.~28, pp. 102--110, 2014.

\bibitem{takashima2015}
Y.~Takashima, T.~Nakashika, T.~Takiguchi, and Y.~Ariki, ``Feature extraction
  using pre-trained convolutive bottleneck nets for dysarthric speech
  recognition,'' in \emph{Proc. EUSIPCO}, 2015, pp. 1426--1430.

\bibitem{lee2016}
T.~Lee, Y.~Liu, P.-W. Huang, J.-T. Chien, W.~K. Lam, Y.~T. Yeung, T.~K.~T. Law,
  K.~Y. Lee, A.~P.-H. Kong, and S.-P. Law, ``Automatic speech recognition for
  acoustical analysis and assessment of {Cantonese} pathological voice and
  speech,'' in \emph{Proc. ICASSP}, 2016, pp. 6475--6479.

\bibitem{yilmaz2016_5}
E.~Y{\i}lmaz, M.~Ganzeboom, C.~Cucchiarini, and H.~Strik, ``Combining
  non-pathological data of different language varieties to improve {DNN-HMM}
  performance on pathological speech,'' in \emph{Proc. INTERSPEECH}, Sept.
  2016, pp. 218--222.

\bibitem{ganzeboom2016_2}
M.~Ganzeboom, E.~Y{\i}lmaz, C.~Cucchiarini, and H.~Strik, ``On the development
  of an {ASR}-based multimedia game for speech therapy: Preliminary results,''
  in \emph{Proc. Workshop MM Health}, Oct. 2016, pp. 3--8.

\bibitem{yilmaz2016_2}
E.~Y{\i}lmaz, M.~Ganzeboom, L.~Beijer, C.~Cucchiarini, and H.~Strik, ``A
  {Dutch} dysarthric speech database for individualized speech therapy
  research,'' in \emph{Proc. LREC}, 2016, pp. 792--795.

\bibitem{yilmaz2017}
E.~Y{\i}lmaz, M.~Ganzeboom, C.~Cucchiarini, and H.~Strik, ``Multi-stage {DNN}
  training for automatic recognition of dysarthric speech,'' in \emph{Proc.
  INTERSPEECH}, Sept. 2017, pp. 2685--2689.

\bibitem{zlokarnik1995}
I.~Zlokarnik, ``Adding articulatory features to acoustic features for automatic
  speech recognition,'' \emph{J.~Acoust.~Soc.~Am.}, vol. 97 (5), p. 3246, 1995.

\bibitem{wrench2000}
A.~A. Wrench and K.~Richmond, ``Continuous speech recognition using
  articulatory data,'' in \emph{Proc. of the International Conference on Spoken
  Language Processing}, 2000, pp. 145--148.

\bibitem{stephenson2000}
T.~A. Stephenson, H.~Bourlard, S.~Bengio, and A.~C. Morris, ``Automatic speech
  recognition using dynamic {Bayesian} networks with both acoustic and
  articulatory variables,'' in \emph{Proc. of the International Conference on
  Spoken Language Processing}, 2000, pp. 951--954.

\bibitem{markov2006}
K.~Markov, J.~Dang, and S.~Nakamura, ``Integration of articulatory and spectrum
  features based on the hybrid {HMM/BN} modeling framework,'' \emph{Speech
  Communication}, vol.~48, no.~2, pp. 161 -- 175, 2006.

\bibitem{mitra2012}
V.~Mitra, N.~H., C.~Espy-Wilson, E.~Saltzman, and L.~Goldstein, ``Recognizing
  articulatory gestures from speech for robust speech recognition,''
  \emph{J.~Acoust.~Soc.~Am.}, vol. 131 (3), p. 2270, 2012.

\bibitem{badino2016}
L.~Badino, C.~Canevari, L.~Fadiga, and G.~Metta, ``Integrating articulatory
  data in deep neural network-based acoustic modeling,'' \emph{Computer Speech
  \& Language}, vol.~36, pp. 173 -- 195, 2016.

\bibitem{rudzicz2011}
F.~Rudzicz, ``Articulatory knowledge in the recognition of dysarthric speech,''
  \emph{IEEE Transactions on Audio, Speech, and Language Processing}, vol.~19,
  no.~4, pp. 947--960, May 2011.

\bibitem{mitra2017}
V.~Mitra, G.~Sivaraman, C.~Bartels, H.~Nam, W.~Wang, C.~Espy-Wilson,
  D.~Vergyri, and H.~Franco, ``Joint modeling of articulatory and acoustic
  spaces for continuous speech recognition tasks,'' in \emph{Proc. ICASSP},
  March 2017, pp. 5205--5209.

\bibitem{mitra2017_2}
V.~Mitra, G.~Sivaraman, H.~Nam, C.~Espy-Wilson, E.~Saltzman, and M.~Tiede,
  ``Hybrid convolutional neural networks for articulatory and acoustic
  information based speech recognition,'' \emph{Speech Communication}, vol.~89,
  pp. 103 -- 112, 2017.

\bibitem{cgn}
N.~Oostdijk, ``The spoken {Dutch} corpus: {Overview} and first evaluation,'' in
  \emph{Proc. LREC}, 2000, pp. 886--894.

\bibitem{jasmin}
C.~Cucchiarini, J.~Driesen, H.~Van~hamme, and E.~Sanders, ``Recording speech of
  children, non-natives and elderly people for {HLT} applications: the
  {JASMIN-CGN Corpus},'' in \emph{Proc. LREC}, May 2008, pp. 1445--1450.

\bibitem{middagphd}
C.~Middag, ``Automatic analysis of pathological speech,'' Ph.D. dissertation,
  Ghent University, Belgium, 2012.

\bibitem{mitra2011}
V.~Mitra, H.~Nam, C.~Y. Espy-Wilson, E.~Saltzman, and L.~Goldstein,
  ``Articulatory information for noise robust speech recognition,'' \emph{IEEE
  Transactions on Audio, Speech, and Language Processing}, vol.~19, no.~7, pp.
  1913--1924, Sept 2011.

\bibitem{browman1992}
B.~C. P. and G.~L., ``Articulatory phonology: an overview.'' \emph{Phonetica},
  vol.~49, no. 3-4, pp. 155--180, 1992.

\bibitem{richmond2001}
K.~Richmond, ``Estimating articulatory parameters from the acoustic speech
  signal,'' Ph.D. dissertation, University of Edinburgh, UK, 2001.

\bibitem{vikramphd}
V.~Mitra, ``Articulatory information for robust speech recognition,'' Ph.D.
  dissertation, University of Maryland, College Park, 2010.

\bibitem{nam2004}
H.~Nam and L.~Goldstein, ``{TADA}: An enhanced, portable task dynamics model in
  {MATLAB},'' \emph{J.~Acoust.~Soc.~Am.}, vol. 115, p. 2430, 2004.

\bibitem{manson2002}
H.~M. Hanson and K.~N. Stevens, ``A quasiarticulatory approach to controlling
  acoustic source parameters in a {Klatt-type} formant synthesizer using
  {HLsyn},'' \emph{J.~Acoust.~Soc.~Am.}, vol. 112, p. 1158, 2002.

\bibitem{mitra2014}
V.~Mitra, G.~Sivaraman, H.~Nam, C.~Espy-Wilson, and E.~Saltzman, ``Articulatory
  features from deep neural networks and their role in speech recognition,'' in
  \emph{Proc. ICASSP}, May 2014, pp. 3017--3021.

\bibitem{collins1996}
B.~Collins and I.~Mees, \emph{The Phonetics of English and Dutch}.\hskip 1em
  plus 0.5em minus 0.4em\relax Koninklijke Brill NV, 1996.

\bibitem{mitra2015}
V.~Mitra and H.~Franco, ``Time-frequency convolutional networks for robust
  speech recognition,'' in \emph{Proc. ASRU}, Dec 2015, pp. 317--323.

\bibitem{kaldi}
D.~Povey, A.~Ghoshal, G.~Boulianne, L.~Burget, O.~Glembek, N.~Goel,
  M.~Hannemann, P.~Motlicek, Y.~Qian, P.~Schwarz, J.~Silovsky, G.~Stemmer, and
  K.~Vesely, ``The {Kaldi} speech recognition toolkit,'' in \emph{Proc. ASRU},
  Dec. 2011.

\end{thebibliography}

\end{document}